\documentclass{article} %
\usepackage{iclr2026_conference,times}

\usepackage{amsmath,amsfonts,bm}

\def\eqref#1{equation~\ref{#1}}

\def\1{\bm{1}}

\DeclareMathAlphabet{\mathsfit}{\encodingdefault}{\sfdefault}{m}{sl}
\SetMathAlphabet{\mathsfit}{bold}{\encodingdefault}{\sfdefault}{bx}{n}

\usepackage{hyperref}
\usepackage{url}
\usepackage{graphicx}

\title{Do Large Language Models Mentalize When They Teach?}

\author{Sevan K. Harootonian\\
Department of Psychology\\
Princeton University\\
\And
Mark K. Ho \\
Department of Psychology \\
New York University \\
\And
Thomas L. Griffiths \\
Department of Computer Science\\
Department of Psychology\\
Princeton University \\
\AND
Yael Niv\thanks{Senior authors contributed equally.} \\
Department of Psychology  \\
Princeton Neuroscience\\
Princeton University \\
\And
Ilia Sucholutsky\footnotemark[1] \\
Department of Psychology \\
New York University\\
}

\iclrfinalcopy %
\begin{document}

\maketitle

\begin{abstract}
How do LLMs decide what to teach next: by reasoning about a learner’s knowledge, or by using simpler rules of thumb? We test this in a controlled task previously used to study human teaching strategies. On each trial, a teacher LLM sees a hypothetical learner’s trajectory through a reward-annotated directed graph and must reveal a single edge so the learner would choose a better path if they replanned. We run a range of LLMs as simulated teachers and fit their trial-by-trial choices with the same cognitive models used for humans: a Bayes-Optimal teacher that infers which transitions the learner is missing (inverse planning), weaker Bayesian variants, heuristic baselines (e.g., reward based), and non-mentalizing utility models. In a baseline experiment matched to the stimuli presented to human subjects, most LLMs perform well, show little change in strategy over trials, and their graph-by-graph performance is similar to that of humans. Model comparison (BIC) shows that Bayes-Optimal teaching best explains most models’ choices. When given a scaffolding intervention, models follow auxiliary inference- or reward-focused prompts, but these scaffolds do not reliably improve later teaching on heuristic-incongruent test graphs and can sometimes reduce performance. Overall, cognitive model fits provide insight into LLM tutoring policies and show that prompt compliance does not guarantee better teaching decisions.
\end{abstract}

\section{Introduction}

Large language models (LLMs) are rapidly being integrated into education as tutors, feedback generators, and teaching assistants. A recent systematic review of 88 empirical studies since ChatGPT’s release reports that intelligent tutoring systems are the most common educational application of LLMs, with overall evidence for improved performance and engagement alongside concerns about over-reliance and reliability \citep{shi2026llmedu}. Complementing these broad surveys, controlled evaluations have begun to test instructional efficacy directly—for example, comparing LLM-generated hints to human-authored help in mathematics \citep{pardos2024plos}, and evaluating AI tutoring systems in randomized trials \citep{kestin2025aitutor}. Parallel work has focused on eliciting better tutoring behavior via design and prompting. Recent approaches propose pedagogical steering methods that prompt an LLM to follow predefined multi-turn tutoring plans \citep[e.g., Productive Failure;][]{puech2025steering}, and tutor-facing copilots that use LLMs to support human tutors in real time and encourage more effective pedagogical moves \citep{wang2025tutorcopilot}. While these lines of research show that LLM-mediated instruction can improve learning outcomes, they primarily benchmark \emph{what} the model produces (and its downstream effects) rather than \emph{how} it chooses instructional actions. Consequently, it remains unclear what teaching strategy an LLM uses when deciding what to teach next.

Work on human teaching provides a useful framework for posing this question. In controlled teaching tasks, people often rely on either (i) a \emph{model-based} strategy that mentalizes a learner’s knowledge and chooses information to maximally improve it, or (ii) \emph{model-free} heuristics that ignore a learner model and rely on environmental cues \citep{harootonian2025mentalizing}. Related contrasts appear in work on adaptive versus routine expertise, and learners penalize instruction that appears untailored \citep{hatano1984two,bass2024teaching}. If LLMs default to heuristics, deployment as tutors may require stronger external scaffolds and monitoring; if they behave in a way that is more model-based, we can better leverage and test that capacity.

Here, we probe LLM teaching strategies by adapting the Graph Teaching task of \citet{harootonian2025mentalizing}. On each trial, an LLM teacher observes a hypothetical learner’s trajectory in a reward-annotated graph and reveals a single edge that could change the learner’s future path. We evaluate multiple LLMs as simulated teachers and fit their trial-by-trial decisions with the same family of cognitive models used in the human study—Bayesian teacher models, heuristic baselines, and non-mentalizing utility models. Finally, we test whether lightweight scaffolding prompts shift LLM teaching behavior in ways analogous to interventions known to affect strategy use in humans.

\section{Methods}

\subsection{Task: Graph Teaching}
\begin{figure}[h]
\begin{center}
\includegraphics[width=.6\textwidth]{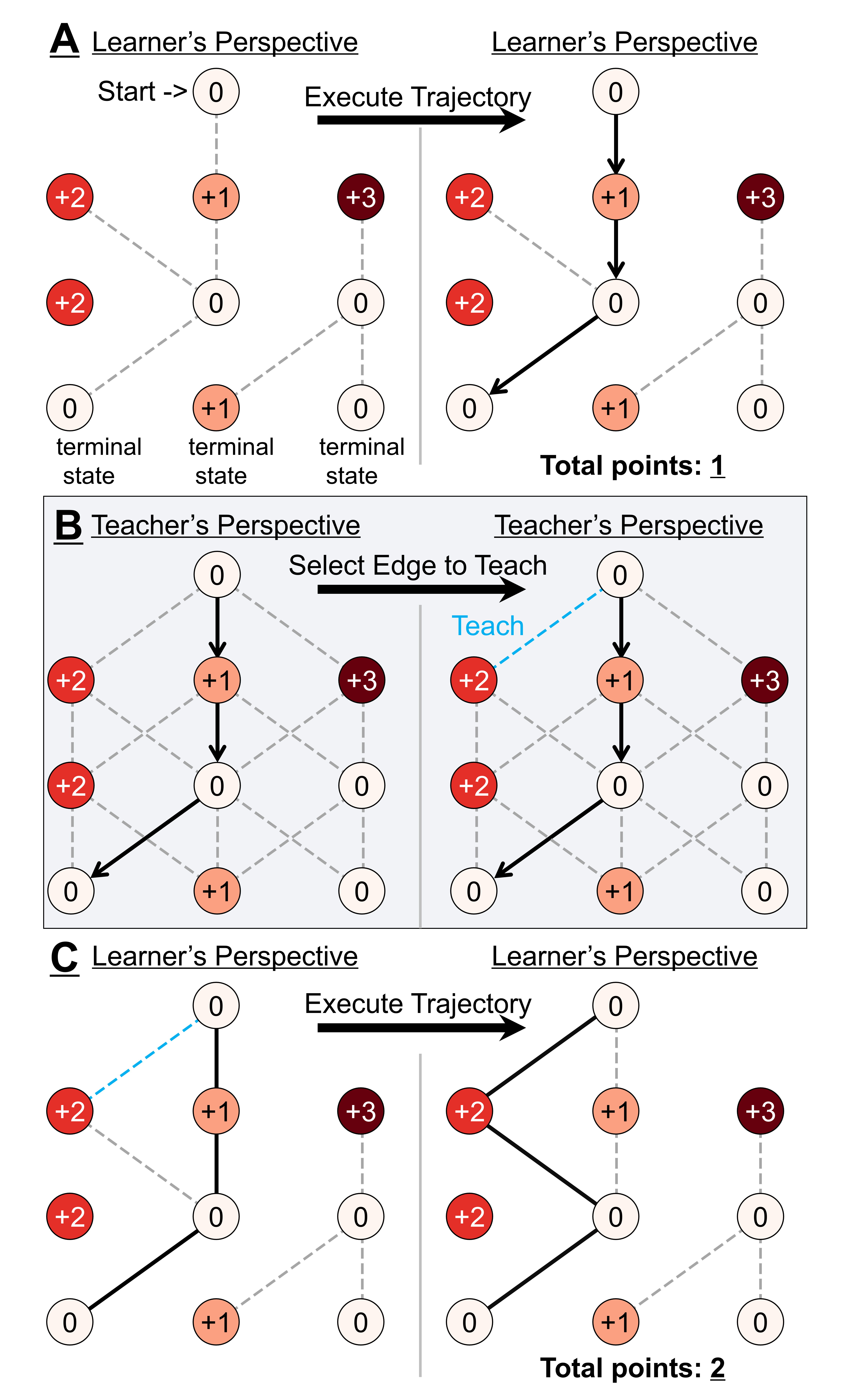}
\end{center}
\vspace{-6mm}
\caption{Graph Teaching Task.
The numeric value in each node indicates the reward obtained when the learner visits that node. The learner starts at the top node and moves down or diagonally down until reaching a terminal node. A)~Learner computes and executes the best available trajectory. B)~Teacher observes the learner's trajectory and selects an edge to teach. C)~Learner incorporates the taught edge and computes and executes a new, better trajectory. Note: participants and models only see part B and make assumptions for parts A and C.}
\label{fig1}
\end{figure}

We used the Graph Teaching task introduced in \citet{harootonian2025mentalizing}.
On each trial, the environment is a directed deterministic acyclic graph with a single start node at the top and multiple terminal nodes at the bottom. Each node carries a non-negative reward, and a learner traverses the graph from top to bottom, moving only straight down or diagonally down, earning the sum of rewards along its path. The learner is assumed to know only an unknown subset of the true edges and to choose a path that is optimal given its current (partial) transition knowledge. The teacher, by contrast, observes the full transition structure and node rewards, as well as a single trajectory taken by the learner on that trial (Figure~\ref{fig1}A).

The teacher knows that the learner is doing the best they can, and must select a single edge to reveal given only the learner’s trajectory, so that if the learner were to replan optimally given this new information, its expected future reward would be as high as possible (Figure~\ref{fig1}B). As in experiments on human participants, unbeknownst to the teacher, the learner is an RL agent that solves for the optimal policy on each trial, given the reward function and subset of the transition matrix. A Full derivation of the learner and task structure can be found in \citet{harootonian2025mentalizing}. In addition, a demo of the human task with full instructions can be found at 
\url{https://sharootonian.github.io/CognitiveStrategiesInTeaching/GraphTeachingTask_demo/}

\subsection{Cognitive model fitting and comparisons}
To characterize the teaching strategies expressed by each LLM, we fit the same family of cognitive models used in the human study: Bayesian Teacher models (which infer a learner knowledge state and maximize expected teaching utility), heuristic models (which rely on task features rather than a learner model), and non-mentalizing utility models (which treat the task as a puzzle-solving problem). Full derivations are provided in \citet{harootonian2025mentalizing}.

Across models, each candidate edge $e$ is assigned a model-specific utility $U_{\text{model}}(e)$ and converted to a choice probability via a softmax rule,
\begin{equation}
P(e) = \frac{\exp\!\big(\beta\,U_{\text{model}}(e)\big)}{\sum_{e'} \exp\!\big(\beta\,U_{\text{model}}(e')\big)},
\end{equation}
where $\beta$ is an inverse-temperature parameter fit to each simulated subject.

\paragraph{Bayesian Teacher models.}
In these models, we model the learner as an agent navigating a directed deterministic graph with transition knowledge $T_L \subseteq T$ and reward function $R$. Given an observed learner trajectory $\zeta$, the Bayes Optimal Teacher (BOT) performs inverse planning by inverting a forward planning model \citep{baker2009action} to infer a posterior over $T_L$ given the observed trajectory $P_{\mathrm{BOT}}(T_L \mid \zeta)$, and evaluates the expected teaching utility of revealing an edge $x$ as
\begin{equation}
U_{\mathrm{BOT}}(x;\zeta)=\sum_{T_L} P_{\mathrm{BOT}}(T_L \mid \zeta)\Big(V_{T_L\cup\{x\}}(s_0)-V_{T_L}(s_0)\Big),
\end{equation}
where $V_{T_L}(s_0)$ is the learner's optimal value at the start state under knowledge $T_L$. We also fit two inference variants that weaken the mentalizing step: the \textbf{No Inverse Planning Bayesian Teacher} replaces inverse planning with a behavioral feasibility likelihood that assigns equal likelihood to all $T_L$ that make $\zeta$ possible, regardless of whether it was optimal, and the \textbf{Prior-only Bayesian Teacher} ignores $\zeta$ and uses only the (flat) prior over $T_L$. In all Bayesian variants, edge utilities are converted to choice probabilities with the same softmax choice rule above.

\paragraph{Heuristic models.}
Our primary heuristic baseline is the \textbf{Reward Heuristic}, which scores each edge $e=[s_i,s_j]$ by the sum of endpoint rewards,
\begin{equation}
F_{\mathrm{reward}}(e)=R(s_i)+R(s_j).
\end{equation}
We additionally include a Depth heuristic based on an edge's vertical position in the graph (with lower edges having higher utility) and a combined Reward+Depth model that learns a linear weighting of the two features via multinomial logistic regression.

\paragraph{Non-mentalizing utility models.}
Finally, we include two models that compute edge utility without representing a learner: a tabular Q-value model (assigning each edge a value from optimal state--action values) and a path-averaged utility model (assigning each edge the mean return over all complete paths containing it).

\paragraph{Model fitting and comparison.}
For each LLM teacher and candidate model, we fit free parameters of the model by maximum likelihood (e.g., $\beta$ for single-utility models and feature weights for multi-feature models) and compare models using Bayesian Information Criterion scores \citep{schwarz1978estimating} (lower values indicate better fit), following the procedure in \citet{harootonian2025mentalizing}.

\subsection{Experimental design}
We match the stimulus sets and trial structures of human Experiments~1 and~3 in \citet{harootonian2025mentalizing}, omitting Experiment~2 (which is conceptually subsumed by Experiments 3's no scaffolding condition).

\paragraph{Simulated teachers.}
We treat each independent run of a given LLM on all the graphs of one of the experiments as a simulated teacher. For each model, we simulate multiple teachers (typically around 30-40 per model in the Baseline Experiment and around 20-30 per condition $\times$ training-group cell in the Scaffolding Intervention Experiment. The conversation context is reset between teachers; within a teacher, the full conversation history is preserved across trials, mirroring a human participant’s ability to remember previous trials. Similar to the human experiment, the LLMs were not given feedback about the outcome of their teaching to prevent learning during the task. 

\paragraph{Baseline Experiment.}
Each simulated teacher in the Baseline Experiment completed 40 trials, with 20 unique graphs and their 20 horizontally
flipped versions presented in random order. These graphs were identical to the orginal graphs used Experiment~1 of the human experiment.

\paragraph{Scaffolding Intervention Experiment.}
The Scaffolding Intervention Experiment followed a $2\times 3$ design crossing training condition (Heuristic Congruent vs.\ Heuristic Incongruent) with scaffolding condition (No Scaffolding, Inference Scaffolding, Reward Scaffolding). On Heuristic-Congruent training trials, graphs were set up such that the Reward Heuristic choice would be the same as the Bayes-Optimal teaching choice, while on the Heuristic-Incongruent training trials, the Reward Heuristic teaching choice was significantly worse than Bayes optimal teaching. The stimulus sets and training/test graphs matched those used in the human Experiment~3: for each training condition, there was a pool of 10 training graphs; all conditions shared the same 5 incongruent test graphs. For each simulated teacher, we construct a sequence of 15 training trials by selecting 5 training graphs from the appropriate pool, adding their flipped versions, and repeating the original 5 (selected $\rightarrow$ flipped $\rightarrow$ selected). We then append the 5 test graphs in a fixed order. Critically, all scaffolding was confined to training trials: test trials in every condition were presented without any auxiliary scaffolding prompt and used only the base teaching task.

\paragraph{Prompts}
At the start of each simulated teacher, we provided task instructions and worked examples that closely matched the human instructions, adapted from the original jsPsych implementation into a text-only description. The instructions explain the graph structure, describe how the learner moves and accumulates rewards, define the teacher’s role, and show examples of good and bad teaching advice.

On each trial, we then presented the teacher with the (full) transition set, node reward values, and the learner’s trajectory, and asked the model to choose a single edge to teach. The trial prompt has the schematic form:
\begin{quote}
\small
\texttt{Try to help this student maximize their points.\\
Teacher's transitions: [list of edges]\\
Reward values: [dictionary of node: reward]\\
Learner's trajectory: [list of edges]\\
Please make sure your response ends in the format (x,y).}
\end{quote}
In the scaffolding conditions of the Scaffolding intervention experiment, training trials included an auxiliary step:
\begin{itemize}
    \item Inference scaffolding: select three edges you think the learner does not know.
    \item Reward scaffolding: select three edges that are connected to the largest-value nodes.
\end{itemize}
Auxiliary responses were requested in the format \texttt{[(x,y),(x,y),(x,y)]}, and the model’s scaffolding response was appended to the conversation history before the subsequent teaching prompt on that trial.
Test trials did not include the auxiliary scaffolding prompt; they used only the base teaching prompt shown above.
The full instruction text and all prompt templates used in our experiments are available in the accompanying code repository.

\section{Results}
We evaluate whether LLMs can act as effective teachers in the Graph Teaching task and whether their teaching behavior is aligned with the cognitive strategies used by humans performing this task, mainly the model-based mentalizing teaching strategy (Bayes Optimal Teacher) or model-free heuristic teaching strategies (e.g., the Reward Heuristic). 

\subsection{Baseline Experiment}

We first examined how teaching performance changes across 40 trials without scaffolding. Because the task provides no feedback and each trial involves a distinct learner, we expected minimal within-experiment learning, as was also seen for human behavior on the task. Consistent with this, Teaching Scores were essentially flat over trials for all model groups: the correlation between trial number and Teaching Score was small in magnitude for every group ($|r|<0.1$; Appendix Fig.~\ref{fig:exp1_learning}). 

\begin{figure}[h]
\begin{center}
\includegraphics[width=1\textwidth]{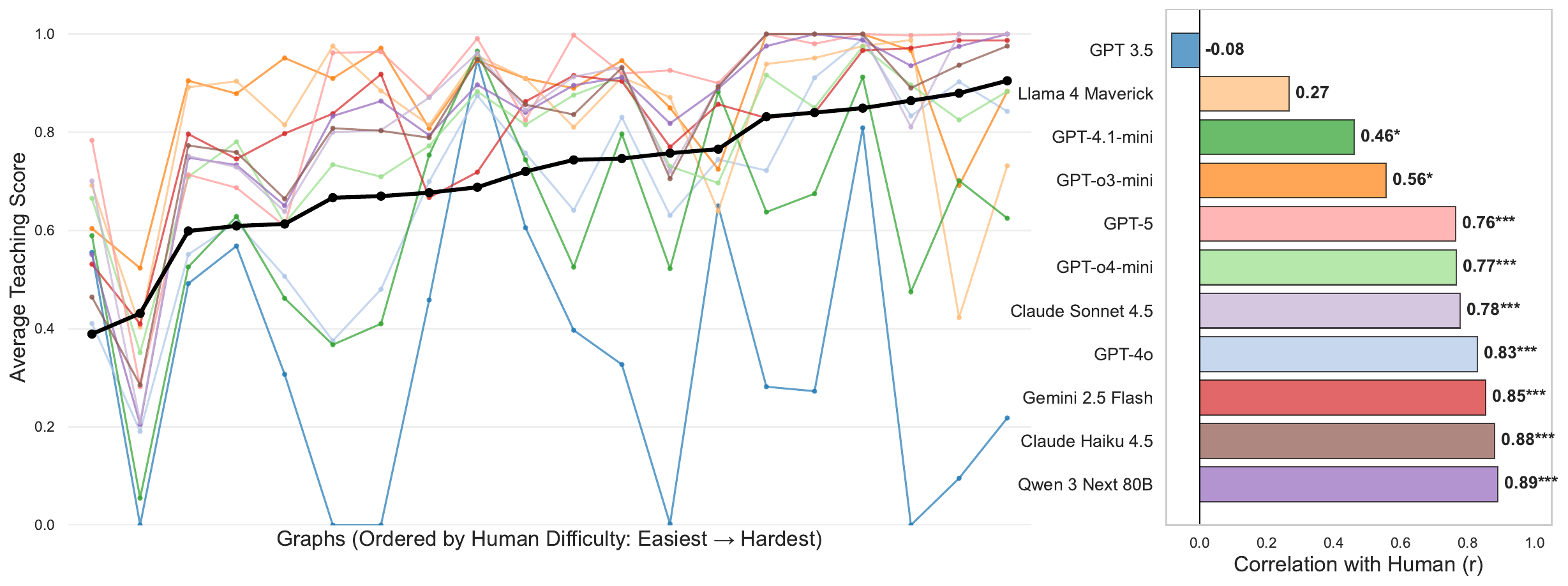}
\end{center}
\vspace{-13px}
\caption{\textbf{Graph-wise performance profiles align with humans.}
Left: mean Teaching Score across the 20 unique graphs configurations for humans (black) and each LLM model. Right: Pearson correlation between each model’s graph-wise performance profile and the human profile; correlations above 0.7 indicate strong alignment with human graph-wise performance ordering. Model colors on the left are the same as on the right. *$p<.05$, ***$p<10^{-4}$}
\label{fig:seed_corrB}
\end{figure}

We next asked whether LLM performance varies across graph instances in a way that mirrors human behavior. Figure~\ref{fig:seed_corrB} shows the mean Teaching Score for each of the 20 unique graph configurations, ordered by their average difficulty for human subjects (left: more difficult, hence lower teaching score). Most models showed strong alignment with humans: the majority exhibited large positive Pearson correlations with human performance (seven models with $r \approx 0.76$--$0.89$, all $p<10^{-4}$), and two additional models showed moderate correlations ($r \approx 0.46$--$0.56$, $p<.05$), suggesting that most of these models are sensitive to different graph properties. GPT-3.5 and Llama-4 Maverick were not significantly correlated with humans (Figure~\ref{fig:seed_corrB}). Overall, these results indicate that for most LLMs, trial-to-trial performance is stable, yet systematically modulated by graph structure in a way that resembles human performance profiles.

\begin{figure}[h]
\begin{center}
\includegraphics[width=1\textwidth]{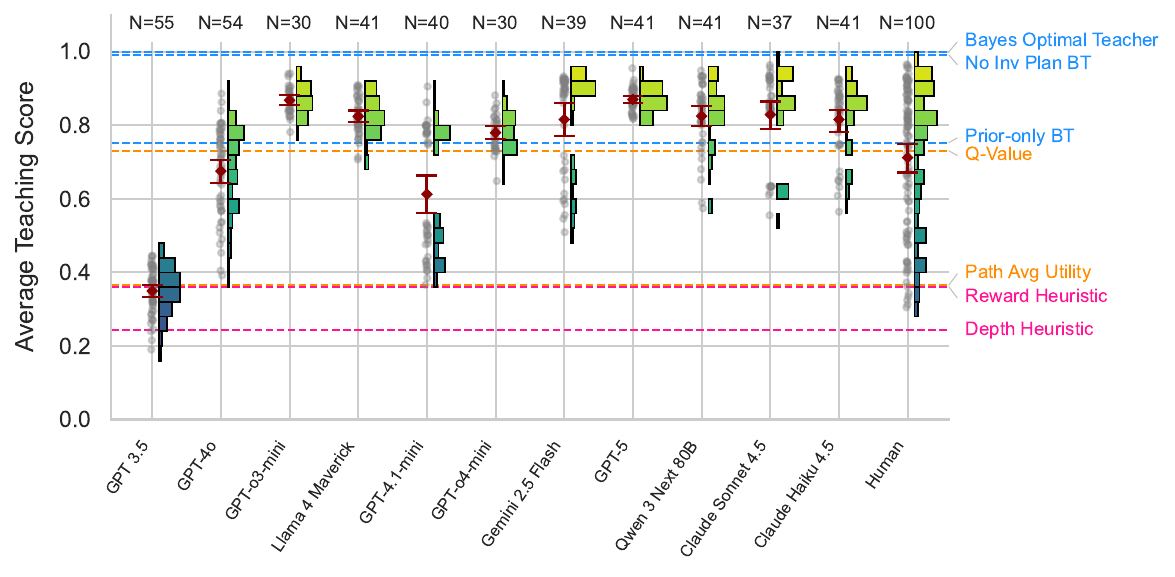}
\end{center}
\vspace{-13px}
\caption{\textbf{Distribution of Teaching Scores.}
Individual-level average Teaching Score in the Baseline Experiment for each LLM model (left) and human subjects (right). Horizontal reference lines show benchmark scores of cognitive models under an argmax policy.}
\label{fig:raincloud}
\end{figure}

We next examined the distribution of individual-level average Teaching Scores (Figure~\ref{fig:raincloud}). Human teaching performance is bimodal, consistent with a mixture of higher-performing mentalizing subjects and lower-performing heuristic subjects. In contrast, most LLM teachers were concentrated in the higher-performing range, close to the Bayes Optimal Teacher benchmark and overlapping with the higher-performing human subjects, although some models showed more variability in Teaching Scores, most notably GPT-o4-mini, Gemini 2.5 Flash and Claude Sonnet 4.5. This suggests greater heterogeneity in their teaching strategies. Among all models, GPT-4o matched human teachers most closely in both overall performance level and variability.

\begin{figure}[h]
\begin{center}
\includegraphics[width=1\textwidth]{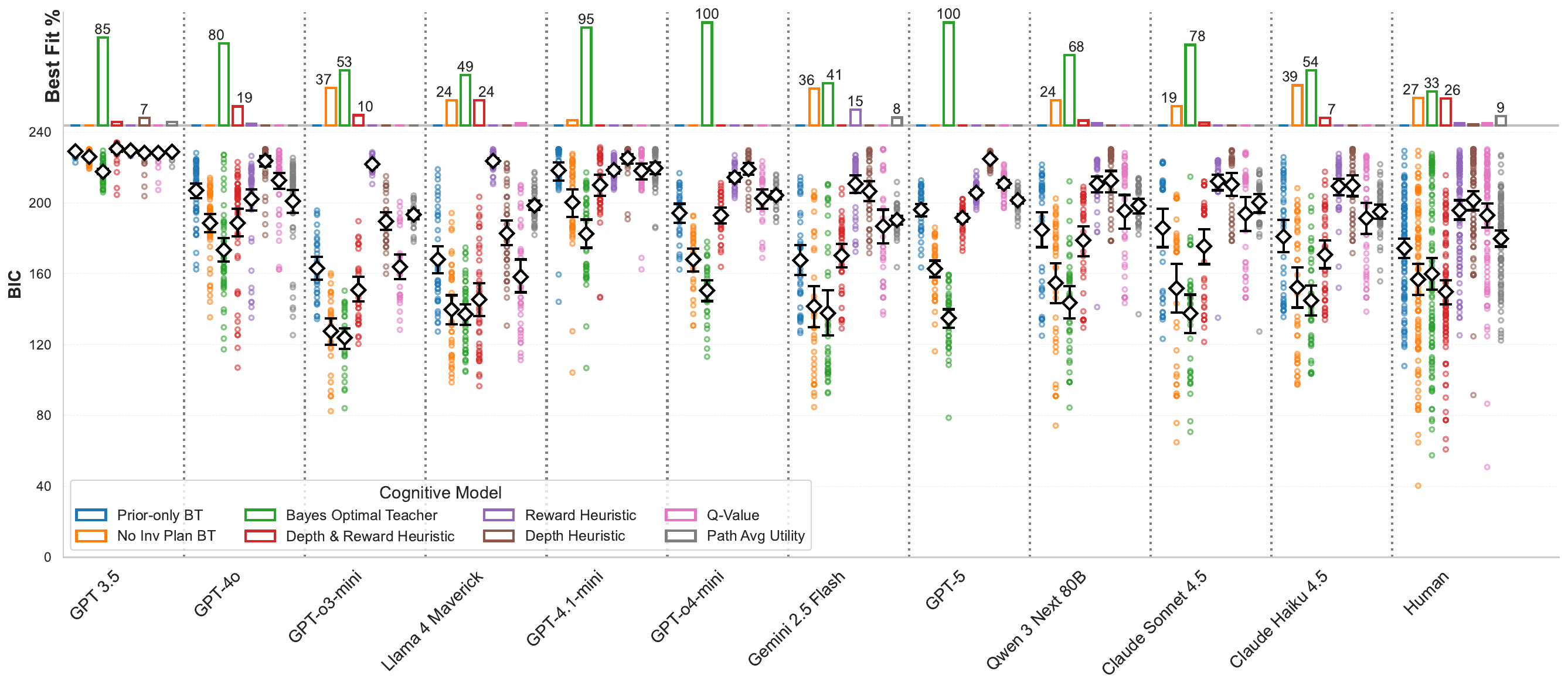}
\end{center}
\vspace{-13px}
\caption{\textbf{Cognitive model fits in the Baseline Experiment.}
BIC scores across candidate teaching models (lower is better), with bars indicating the fraction of simulated teachers best fit by each cognitive model (left) and for human subjects (right).}
\label{fig:bic_exp1}
\end{figure}

To characterize strategies more directly, we fit the cognitive models to each simulated teacher’s trial-by-trial choices and compared models using BIC. Figure~\ref{fig:bic_exp1} shows that the Bayes Optimal Teacher provides the best overall account for the behavior of most models, consistent with their high Teaching Scores. Relative to the human data, LLMs showed substantially less alignment with heuristic and non-mentalizing utility strategies, even among lower-performing LLM teachers. The clearest heterogeneity appears for GPT-o3-mini, Llama 4 Maverick, Gemini 2.5 Flash, and Claude Haiku 4.5, which show more mixed distributions of best-fitting models similar to humans.

\subsection{Scaffolding Intervention Experiment}
We next asked whether the same intervention that shifted human teaching strategies \citep{harootonian2025mentalizing} can also shape LLM teaching. This experiment added two manipulations. First, we varied the training environment: in the Heuristic Congruent condition, a simple Reward Heuristic tended to point to the same edge as the Bayes-Optimal Teacher during training trials, whereas in the Heuristic Incongruent condition the Reward Heuristic was systematically misleading, so high performance required reasoning about the learner’s knowledge. Second, we varied whether training trials included an auxiliary scaffolding step before teaching: No Scaffolding (teach directly, as in the Baseline Experiment), Inference Scaffolding (mark three edges the learner may not know, then teach), or Reward Scaffolding (mark three edges connected to high-value nodes, then teach). All groups then completed the same set of heuristic-incongruent test trials without any auxiliary task.

In the human experiment, Heuristic Congruent training increased reliance on the Reward Heuristic and led to worse test performance than Heuristic Incongruent training in the No-Scaffolding group. Inference scaffolding improved teaching performance relative to No Scaffolding, and this benefit persisted into the test phase even after the auxiliary scaffolding prompt was removed. Reward Scaffolding, in contrast, increased heuristic-consistent teaching and could reduce performance on incongruent test trials. Both scaffolding manipulations also removed the congruent--incongruent training difference seen in No Scaffolding \citep{harootonian2025mentalizing}.

We applied the same $2\times 3$ design to analyses of LLM performance, %
asking whether scaffolding would increase LLM Teaching Scores at test and shift model fits toward Bayes Optimal Teacher, as in humans.

\begin{figure}[h]
\vspace{-5px}
\begin{center}
\includegraphics[width=1\textwidth]{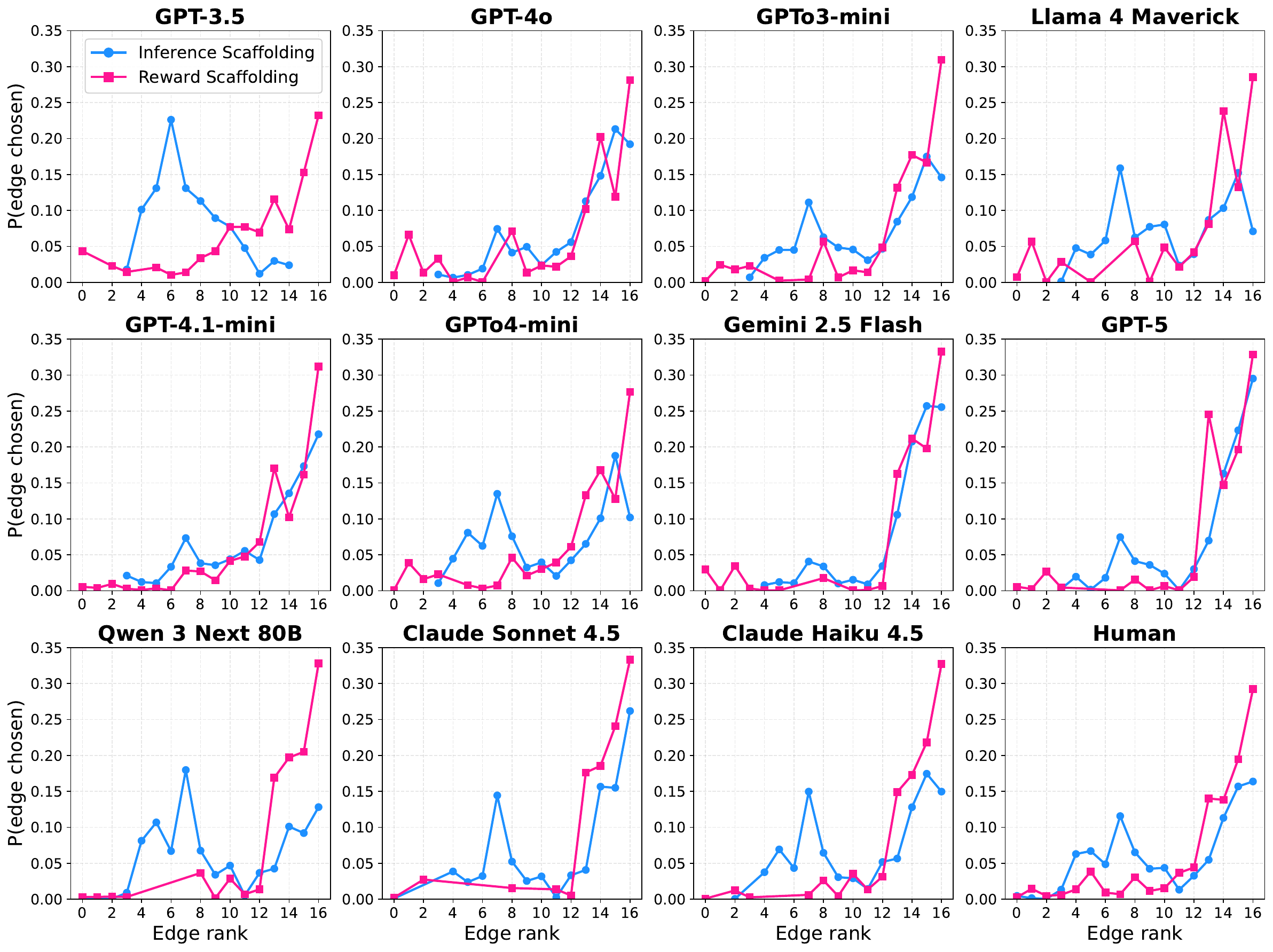}
\end{center}
\vspace{-13px}
\caption{\textbf{Auxiliary scaffolded edge selections.} Blue line shows probabilities of choosing edges in the inference scaffolding condition, with edges ordered according to the predicted probability that an edge is unknown to the learner according to the Bayes-Optimal Teacher (0 - most likely known, 16 - most likely unknown). Pink shows probabilities of choosing edges in the reward scaffolding condition, with edges ordered according to the value of each edge as a function of the sum of the two nodes it connects, as defined by the Reward Heuristic (0 - lowest value, 16 - highest value). In both cases, values to the right correspond to edges more consistent with the instruction for this condition, but note that the order of edges differs between the Bayes-Optimal Teacher and the Reward Heuristic, and thus for the pink and blue lines. Because three edges were marked on each trial, the y-axis shows the probability that an edge was one of the three selected (maximum: 33.3\%), averaged over all trials and LLM teachers. Most LLM responses aligned with the model predictions, indicating that participants performed the scaffolding task as intended.
}
\label{fig:scaffold_select}
\end{figure}

We first evaluated performance on the auxiliary selection step itself. Figure~\ref{fig:scaffold_select} shows that under Reward Scaffolding they preferentially selected edges ranked highest by reward, and under Inference Scaffolding they showed a systematic preference for edges ranked as more likely unknown to the learner. This indicates that LLMs can reliably execute the scaffolding task and produce structured selections aligned with the human behavior.

\begin{figure}[h]
\begin{center}
\includegraphics[width=1\textwidth]{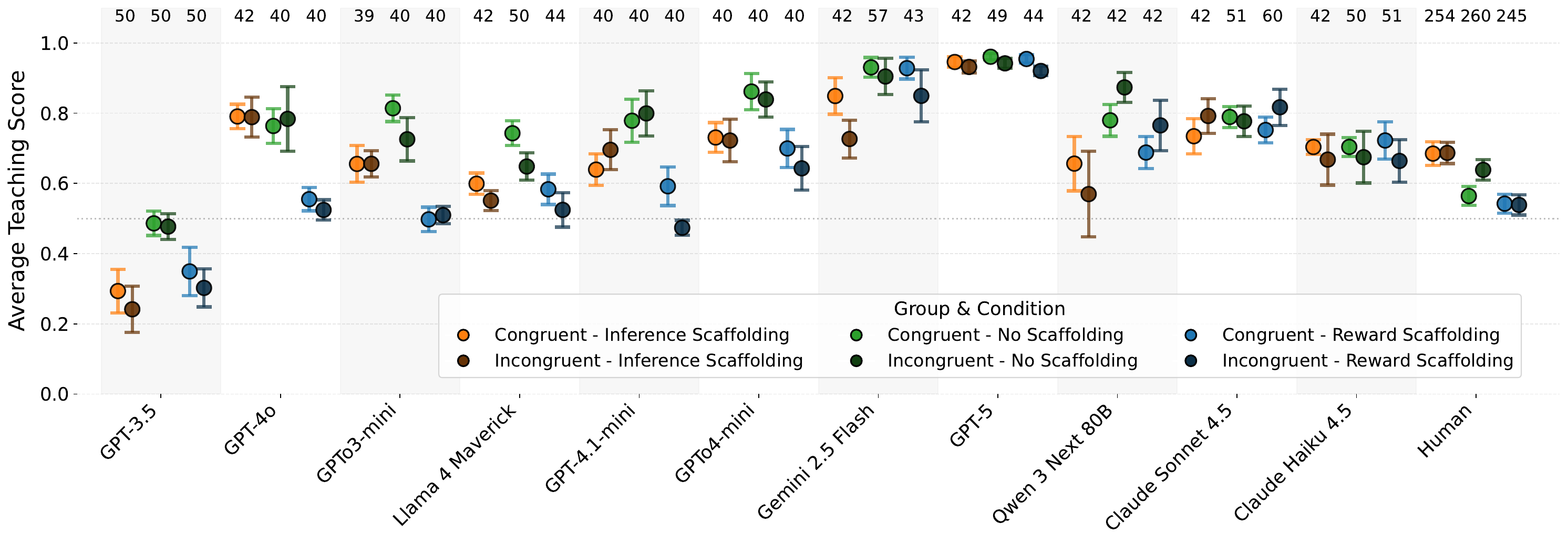}
\end{center}
\vspace{-13px}
\caption{\textbf{Teaching performance across scaffolding and congruency conditions.}
Mean Teaching Score at test for No Scaffolding, Inference Scaffolding, and Reward Scaffolding, shown separately for congruent and incongruent training groups. Error bars show 95\% CI of the mean. Top: number of simulated teachers for each scaffolding condition (across congruent and incongruent training).}
\label{fig:exp3_perf}
\end{figure}

Despite strong auxiliary-step compliance, scaffolding did not translate into improved teaching. Figure~\ref{fig:exp3_perf} shows mean Teaching Score across test trials as a function of scaffolding condition and training congruency. Unlike humans, LLMs showed little evidence of a consistent increase in Teaching Score after Inference Scaffolding (left, orange) relative to No Scaffolding (middle, green) at test. Instead, several models exhibited generally reduced teaching performance after training with scaffolding prompts, most notably under Reward Scaffolding (right, blue). This pattern was seen in some models even during the training trials, where teaching was done immediately after scaffolding (Supp. Figure~\ref{fig:exp3_perf_train}). Moreover, the congruency effect seen in humans was inconsistent in LLMs.

\begin{figure}[h]
\begin{center}
\includegraphics[width=1\textwidth]{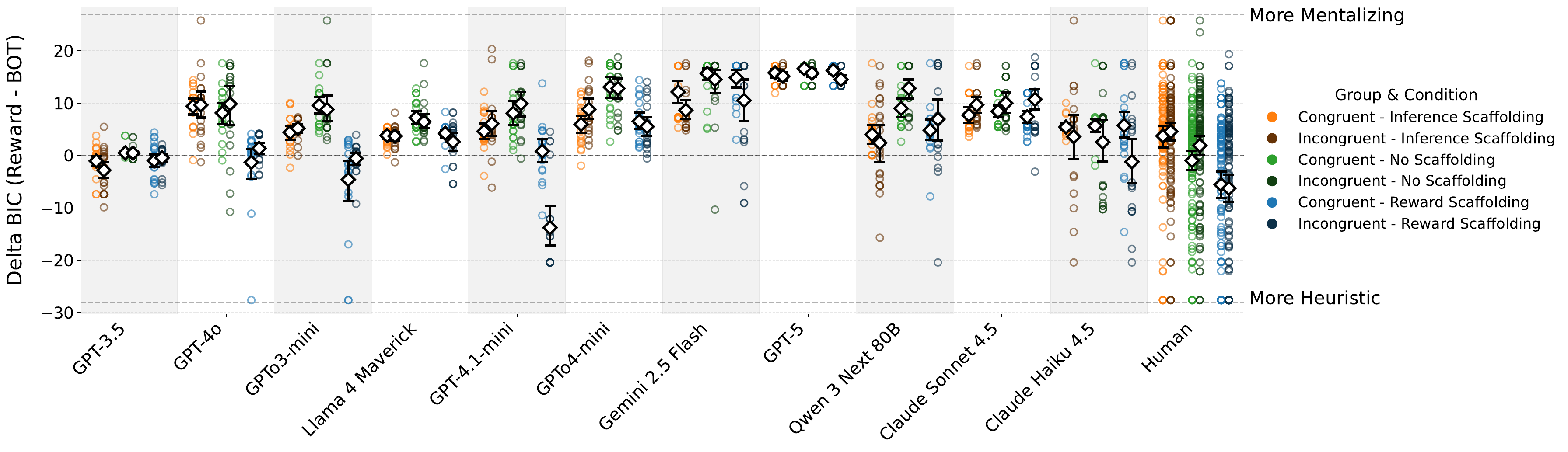}
\end{center}
\vspace{-13px}
\caption{\textbf{Model fits at test across scaffolding conditions.} Difference in BIC between the Reward Heuristic and the Bayes Optimal Teacher ($\Delta\mathrm{BIC} = \mathrm{BIC}_\mathrm{Reward}-\mathrm{BIC}_\mathrm{Bayes\,Optimal\,Teacher}$)  on test trials. Positive values favor the Bayes Optimal Teacher; negative values favor the Reward Heuristic.}
\label{fig:delta_bic}
\end{figure}

Finally, we compared consistency with the Bayes Optimal Teacher and the Reward Heuristic on test trials using $\Delta$BIC (Figure~\ref{fig:delta_bic}). Across models and conditions, the Bayes Optimal Teacher provided a better fit to LLM behavior, consistent with the high Teaching Scores observed in the Baseline Experiment. Reward Scaffolding reduces this advantage for a subset of models, making the Reward Heuristic relatively more competitive, but we did not observe a broad shift toward heuristic-dominant fits analogous to that seen in humans. Together, these results show a dissociation: LLMs can perform the auxiliary scaffolding step in the intended way, but the scaffolding intervention that improves human teaching does not reliably improve LLM teaching performance and can sometimes impair it.

\section{Discussion}

Across both experiments, we find that most LLMs achieve high Teaching Scores and are best fit by the Bayes Optimal Teacher. Unlike human teachers, the scaffolding intervention in the second experiment did not reliably improve LLM teaching performance and sometimes impaired it, even when the scaffolding highlighted the better choice (Inference Scaffolding). These findings suggest that in this structured teaching problem, many contemporary LLMs select instructional actions that are well described by a learner model based (mentalizing) account, and they do not reliably benefit from the same lightweight scaffolds that shift strategy use in humans.

A key result is that LLM performance is stable across trials (consistent with the task providing no feedback and each trial involving a new learner), but varies systematically with graph structure in a way that often mirrors human difficulty profiles. This alignment at the level of item-by-item performance supports the idea that LLMs are sensitive to the same underlying problem features that make teaching harder or easier for people, even though the distribution of strategies differs: whereas human teachers show a prominent mixture of mentalizing and heuristic patterns, most LLMs concentrate in the higher-performing range and are predominantly fit by the Bayes Optimal Teacher.

One interpretation of these differences comes from a resource-rational view of strategy selection.
In humans, mentalizing about a learner can be effective but cognitively costly, creating pressure to rely on simpler, lower-effort heuristics when these are ``good enough'' \citep{lieder2020resource,harootonian2025mentalizing}. By contrast, LLMs may face a very different cost landscape: at inference time, executing a computation that resembles inverse planning may not incur the same subjective effort costs that shape human strategy choice. This makes it less surprising that many models appear more consistently ``model-based'' in our task, especially for reasoning-oriented systems that are optimized to externalize intermediate computations (e.g., via chain-of-thought style reasoning) \citep{wei2022chain}. At the same time, the resource-rational perspective highlights a concrete direction for future work: if strategy selection in LLMs is cost-sensitive, then manipulating test-time constraints (latency, context length, or explicit token budgets) should produce predictable shifts toward simpler teaching rules, paralleling the way human strategy use changes under resource constraints.

The scaffolding results sharpen this comparison. LLMs generally followed the auxiliary instructions (showing they can perform the inference- or reward-oriented marking step), yet this compliance did not translate into reliably better subsequent teaching choices after Inference Scaffolding and worse performance after Reward Scaffolding, as seen in humans. This dissociation suggests that ``completing the scaffolded step'' and using it to guide action selection are separable in LLMs: the auxiliary step may function more like an additional instruction-following subtask than like a cognitive support that reduces internal costs or changes the effective decision policy. More practically, this finding cautions that prompting-based scaffolds for tutoring (even when they produce sensible intermediate outputs) may not improve pedagogical decisions unless the scaffold is tightly coupled to the model's action-selection process.

A further caveat is that strong fits to a mentalizing model do not guarantee that LLMs are performing human-like mental state inference. Because heuristic policies can be highly effective and can overlap with optimal teaching on many trials, an LLM could implement a sophisticated learned mapping from trajectories and rewards to teaching actions that mimics Bayes-optimal behavior without explicitly representing learner knowledge. This possibility aligns with views in cognitive modeling where heuristics can approximate Bayesian solutions or appear Bayesian under strong priors \citep{parpart2018heuristics}, and where repeated exposure can yield amortized computations that reproduce inference-like behavior with minimal online deliberation \citep{gershman2014amortized,dasgupta2018remembrance}. Distinguishing ``genuine'' learner-model computation from high-capacity heuristic approximation will likely require more diagnostic tests (e.g., adversarial graph families where shortcuts fail, manipulations that decouple reward salience from information gain, or out-of-distribution learner policies that break simple pattern matching).

Taken together, our results suggest that mechanistic evaluations of teaching policy provide a complementary lens to outcome-focused benchmarks of AI tutoring. In settings where LLMs already behave in a model-based manner, adding generic scaffolding prompts may offer limited benefit and can introduce new failure modes. Conversely, if future models or deployment settings induce heuristic shortcuts (e.g., under stronger compute constraints or noisier contexts), then scaffolds that explicitly couple learner-state inference to action selection may be necessary. More broadly, importing cognitive models of teaching offers a principled way to characterize \emph{how} LLM tutors choose what to teach next, and to design interventions that target strategy rather than surface compliance.

\subsubsection*{Acknowledgments}
MKH was supported by NSF Award \#2348442

\bibliography{iclr2026_conference}
\bibliographystyle{iclr2026_conference}

\newpage
\appendix
\setcounter{figure}{0}
\renewcommand{\thefigure}{S\arabic{figure}}

\setcounter{table}{0}
\renewcommand{\thetable}{S\arabic{table}}

\section{Supplementary Material}

\label{app:impl}

\subsection{Teaching score}
We measure trial-level performance using the Teaching Score: the expected teaching utility of the chosen edge, normalized by the maximum possible teaching utility on that trial. Concretely, for trial $t$ with candidate edges $E_t$, we compute
\begin{equation}
\mathrm{TeachingScore}_t = \frac{U_{\mathrm{BOT}}(e_t)}{\max_{e\in E_t} U_{\mathrm{BOT}}(e)},
\end{equation}
where $U_{\mathrm{BOT}}$ is the Bayes Optimal Teacher expected teaching utility (Eq.~2 in \citet{harootonian2025mentalizing}). We use the pre-computed $U_{\mathrm{BOT}}(e)$ values distributed with the original stimulus files, ensuring that scoring is identical to the human experiments.

\subsection{Models and decoding}
We evaluated a set of LLMs via their public APIs: OpenAI models (GPT-5, GPT-4o, GPT-4.1-mini, GPT-3.5, GPT-o1-mini, GPT-o3-mini, GPT-o4-mini), Anthropic models (Claude Sonnet~4.5, Claude Haiku~4.5), Google Gemini~2.5 Flash, and Together-hosted open-weight models (Llama-4 Maverick 17B Instruct, Qwen3-Next 80B Instruct).

For the Anthropic, Gemini, Llama-4, Qwen3, GPT-4o, GPT-4.1-mini, and GPT-3.5 models  we explicitly set the temperature to zero and, where supported, capped the maximum number of generated tokens (typically at 4096) in order to reduce sampling noise. For the OpenAI GPT models we left decoding parameters at the provider defaults and did not explicitly set the temperature or nucleus-sampling parameters. Across all models, we disabled tool use and function calling and requested plain-text replies. A full list of API model identifiers and decoding parameters is provided in Appendix~\ref{app:impl}.

\subsection{Response parsing.}
We extract the advised edge using a regular expression that matches coordinate pairs of the form \texttt{(digit,digit)}. If multiple edges are present (for example, during explanation plus a final answer), we take the last matched pair as the chosen edge. If no valid edge is found, the trial is recorded but treated as missing; we do not re-prompt or retry.

\clearpage
\subsection{Task Instruction Prompt}

Welcome! In this experiment, you will act as a teacher helping a student choose the best path through a graph to maximize points.

\paragraph{Graph Structure:}
\begin{itemize}
    \item The graph consists of circles connected by lines.
    \item Each circle represents a point value.
\end{itemize}

\paragraph{How the Student Navigates:}
\begin{itemize}
    \item The student moves down a graph, gathering points along the way.
    \item They always start at the top and move down or diagonally down to reach an endpoint.
    \item Each circle provides a certain number of points.
    \item The student does not know all possible paths and will choose what they believe to be the best option based on their knowledge.
\end{itemize}

Here is an example of the student's path:

\paragraph{Example Walkthrough:}
\begin{itemize}
    \item The student navigates through the graph, earning points along the way.
    \item They take a path they believe is optimal, given what they know.
    \item In this example, they earn 4 points because they were unaware of a better route.
\end{itemize}

\begin{quote}
\small\ttfamily
Learner's transitions: [(0,1),(0,2),(1,5),(2,4),(2,5),(2,6),(5,8),(5,9),(6,9)]\\
Reward values: \{0:0,1:2,2:1,3:0,4:0,5:1,6:0,7:3,8:1,9:0\}\\
Learner's Trajectory: [(0,1),(1,5),(5,8)]
\end{quote}

\paragraph{Your Role as a Teacher:}
\begin{itemize}
    \item You can see all possible paths.
\end{itemize}

\begin{quote}
\small\ttfamily
Teacher's transitions: [\\
\hspace*{2em}(0,1),(0,2),(0,3),(1,4),(1,5),(2,4),(2,5),(2,6),\\
\hspace*{2em}(3,5),(3,6),(4,7),(4,8),(5,7),(5,8),(5,9),(6,8),(6,9)]
\end{quote}

\begin{itemize}
    \item You do not know the student's transitions, but know the Reward values, Learner's Trajectory and you assume they took the best path available to them.
    \item Your goal is to reveal one new path to the student to improve their score.
\end{itemize}

\paragraph{Example of Good Advice:}
\begin{quote}
\small\ttfamily
Teacher's transitions: [\\
\hspace*{2em}(0,1),(0,2),(0,3),(1,4),(1,5),(2,4),(2,5),(2,6),\\
\hspace*{2em}(3,5),(3,6),(4,7),(4,8),(5,7),(5,8),(5,9),(6,8),(6,9)]\\
Reward values: \{0:0,1:2,2:1,3:0,4:0,5:1,6:0,7:3,8:1,9:0\}\\
Learner's Trajectory: [(0,1),(1,5),(5,8)]\\
If you reveal the edge (5,7) to the student, the student will navigate again and now earn 6 points instead of 4.
\end{quote}

\paragraph{Example of Bad Advice:}
\begin{quote}
\small\ttfamily
Teacher's transitions: [\\
\hspace*{2em}(0,1),(0,2),(0,3),(1,4),(1,5),(2,4),(2,5),(2,6),\\
\hspace*{2em}(3,5),(3,6),(4,7),(4,8),(5,7),(5,8),(5,9),(6,8),(6,9)]\\
Reward values: \{0:0,1:2,2:1,3:0,4:0,5:1,6:0,7:3,8:1,9:0\}\\
Learner's Trajectory: [(0,1),(1,5),(5,8)]\\
If you reveal the edge (5,9) to the student, the student will navigate again and will not increase their points.
\end{quote}

Here are 3 examples of good advice:

\begin{quote}
\small\ttfamily
Teacher's transitions: [\\
\hspace*{2em}(0,1),(0,2),(0,3),(1,4),(1,5),(2,4),(2,5),(2,6),\\
\hspace*{2em}(3,5),(3,6),(4,7),(4,8),(5,7),(5,8),(5,9),(6,8),(6,9)]\\
Reward values: \{0:0,1:2,2:1,3:0,4:0,5:1,6:0,7:2,8:1,9:0\}\\
Learner's Trajectory: [(0,2),(2,5),(5,8)]\\
Advice: (5,7) is the best edge to reveal.
\end{quote}

\begin{quote}
\small\ttfamily
Teacher's transitions: [\\
\hspace*{2em}(0,1),(0,2),(0,3),(1,4),(1,5),(2,4),(2,5),(2,6),\\
\hspace*{2em}(3,5),(3,6),(4,7),(4,8),(5,7),(5,8),(5,9),(6,8),(6,9)]\\
Reward values: \{0:0,1:2,2:1,3:0,4:0,5:1,6:0,7:2,8:1,9:2\}\\
Learner's Trajectory: [(0,2),(2,5),(2,8)]\\
Advice: Here are 2 good edges to reveal: (5,7) or (5,9).
\end{quote}

\begin{quote}
\small\ttfamily
Teacher's transitions: [\\
\hspace*{2em}(0,1),(0,2),(0,3),(1,4),(1,5),(2,4),(2,5),(2,6),\\
\hspace*{2em}(3,5),(3,6),(4,7),(4,8),(5,7),(5,8),(5,9),(6,8),(6,9)]\\
Reward values: \{0:0,1:0,2:0,3:1,4:0,5:2,6:1,7:0,8:2,9:0\}\\
Learner's Trajectory: [(0,3),(3,6),(6,8)]\\
Advice: Here are 2 good edges to reveal: (3,5) or (5,8).
\end{quote}

Okay let's get started!
I will give the Teacher's transitions, Reward values, and the Learner's Trajectory. And you have to select only 1 edge from the Teacher's transitions to reveal to the student.

\subsubsection{Trial prompt (no scaffolding)}
\begin{quote}
\small\ttfamily
Try to help this student maximize their points.\\
Teacher's transitions: \{teacher\_graph\}\\
Reward values: \{learner\_rewards\}\\
Learner's Trajectory: \{traj\}\\
Please make sure your response ends in the format (x,y)
\end{quote}

\subsubsection{Additional instructions for reward scaffolding}
We are going to add a new step before teaching. Instead of teaching an edge you will select three edges that are connecting the largest numbers.

Here are 2 examples of good advice:

\begin{quote}
\small\ttfamily
Teacher's transitions: [\\
\hspace*{2em}(0,1),(0,2),(0,3),(1,4),(1,5),(2,4),(2,5),(2,6),\\
\hspace*{2em}(3,5),(3,6),(4,7),(4,8),(5,7),(5,8),(5,9),(6,8),(6,9)]\\
Reward values: \{0:0,1:3,2:2,3:3,4:1,5:2,6:0,7:0,8:1,9:3\}\\
Learner's Trajectory: [(0,2),(2,4),(4,6)]\\
The three best choices are [(4,8),(2,5),(0,3)]\\
Explanation for (4,8): This edge connects to node 8, which has a higher reward value.\\
Explanation for (2,5): This edge connects to node 5, which has a higher reward value than the student's current path.\\
Explanation for (0,3): This edge connects to nodes with high values that could lead to better outcomes.
\end{quote}

\subsubsection{Reward scaffolding trial prompt (training trials only)}
\begin{quote}
\small\ttfamily
Select three edges that are connected to the largest value nodes.\\
Teacher's transitions: \{teacher\_graph\}\\
Reward values: \{learner\_rewards\}\\
Learner's Trajectory: \{traj\}\\
Please make sure your response ends in the format [(x,y),(x,y),(x,y)]
\end{quote}

After scaffolding response, the teaching trial prompt is given:

\begin{quote}
\small\ttfamily
Try to help this student maximize their points.\\
Teacher's transitions: \{teacher\_graph\}\\
Reward values: \{learner\_rewards\}\\
Learner's Trajectory: \{traj\}\\
Please make sure your response ends in the format (x,y)
\end{quote}

\subsubsection{Additional instructions for inference scaffolding}
We are going to add a new step before teaching. Instead of teaching an edge you will select three edges that you think the student does not know.

Here are 2 examples of good advice:

\begin{quote}
\small\ttfamily
Teacher's transitions: [\\
\hspace*{2em}(0,1),(0,2),(0,3),(1,4),(1,5),(2,4),(2,5),(2,6),\\
\hspace*{2em}(3,5),(3,6),(4,7),(4,8),(5,7),(5,8),(5,9),(6,8),(6,9)]\\
Reward values: \{0:0,1:0,2:1,3:3,4:2,5:1,6:0,7:2,8:2,9:1\}\\
Learner's Trajectory: [(0,2),(2,5),(5,9)]\\
The three best choices are [(5,7),(5,8),(2,4)]\\
Explanation for (5,7) and (5,8): If the learner had known about (5,7) or (5,8), they would have chosen it. Therefore, we can confidently assume the learner does not know (5,7) or (5,8).\\
Explanation for (2,4): There's a higher chance that the learner knows (4,7) or (4,8) rather than (2,4) itself.
\end{quote}

\begin{quote}
\small\ttfamily
Teacher's transitions: [\\
\hspace*{2em}(0,1),(0,2),(0,3),(1,4),(1,5),(2,4),(2,5),(2,6),\\
\hspace*{2em}(3,5),(3,6),(4,7),(4,8),(5,7),(5,8),(5,9),(6,8),(6,9)]\\
Reward values: \{0:0,1:3,2:2,3:3,4:1,5:2,6:0,7:0,8:1,9:3\}\\
Learner's Trajectory: [(0,2),(2,4),(4,6)]\\
The three best choices are [(4,8),(2,5),(0,3)]\\
Explanation for (4,8): If the learner had known about (4,8), they would have chosen it. Therefore, we can confidently assume the learner does not know (4,8).\\
Explanation for (2,5): There's a higher chance that the learner knows one of edges connected to the end of (2,5) rather than (2,5) itself.\\
Explanation for (0,3): This one is a little more complex. We've seen the learner use (4,6). So using the same reasoning as (2,5) we find that (0,3) is most likely not known by the learner.
\end{quote}

\subsubsection{Inference scaffolding trial prompt (training trials only)}
\begin{quote}
\small\ttfamily
Select three edges that you think the student does not know.\\
Teacher's transitions: \{teacher\_graph\}\\
Reward values: \{learner\_rewards\}\\
Learner's Trajectory: \{traj\}\\
Please make sure your response ends in the format [(x,y),(x,y),(x,y)]
\end{quote}

After scaffolding response, the teaching trial prompt is given:

\begin{quote}
\small\ttfamily
Try to help this student maximize their points.\\
Teacher's transitions: \{teacher\_graph\}\\
Reward values: \{learner\_rewards\}\\
Learner's Trajectory: \{traj\}\\
Please make sure your response ends in the format (x,y)
\end{quote}

\clearpage

\subsection{Generation parameters}

All models were tested with the following generation parameters. When a parameter is listed as \emph{default}, it was not explicitly set in the code and the API's default value was used.

\paragraph{OpenAI models.}
\begin{itemize}
    \item \texttt{gpt-5}: temperature default (not specified in code); top\_p default; max\_tokens default; logprobs default; other parameters: none.
    \item \texttt{gpt-4o}: temperature $=0$; top\_p default; max\_tokens default; logprobs enabled (top 5); other parameters: none.
    \item \texttt{gpt-3.5-turbo}: temperature $=0$; top\_p default; max\_tokens default; logprobs enabled (top 5); other parameters: none.
    \item \texttt{gpt-4.1-mini}: temperature $=0$; top\_p default; max\_tokens default; logprobs enabled (top 5); other parameters: none.
    \item \texttt{o1-mini} (model: \texttt{o1-mini}): temperature not supported by this model; reasoning\_effort not specified (default); logprobs not supported by this model; other parameters: none. Note: o1-series models do not support temperature or logprobs parameters.
    \item \texttt{o3-mini} (model: \texttt{o3-mini}): temperature not supported by this model; reasoning\_effort \texttt{high}; logprobs not supported by this model; other parameters: none. Note: o1-series models do not support temperature or logprobs parameters.
    \item \texttt{o4-mini} (model: \texttt{o4-mini}): temperature not supported by this model; reasoning\_effort \texttt{low}; logprobs not supported by this model; other parameters: none. Note: o1-series models do not support temperature or logprobs parameters.
\end{itemize}

\paragraph{Anthropic models.}
\begin{itemize}
    \item \texttt{claude-sonnet-4.5 (claude-sonnet-4-5-20250929)}: temperature $=0$; top\_p default; max\_tokens $=4096$; other parameters: none.
    \item \texttt{claude-haiku-4.5 (claude-haiku-4-5-20250929)}: temperature $=0$; top\_p default; max\_tokens $=4096$; other parameters: none.
\end{itemize}

\paragraph{Google models.}
\begin{itemize}
    \item \texttt{gemini-2.5-flash}: temperature $=0$; top\_p default; max\_output\_tokens default; other parameters: none.
\end{itemize}

\paragraph{Together AI models.}
\begin{itemize}
    \item \texttt{Llama-4-Maverick-17B-128E-Instruct-FP8}: temperature $=0$; top\_p default; max\_tokens $=4096$; other parameters: none.
    \item \texttt{Qwen3-Next-80B-A3B-Instruct}: temperature $=0$; top\_p default; max\_tokens $=4096$; other parameters: none.
\end{itemize}

\subsection{Supplementary Figures}
\clearpage
\begin{figure}
\begin{center}
\includegraphics[width=1\textwidth]{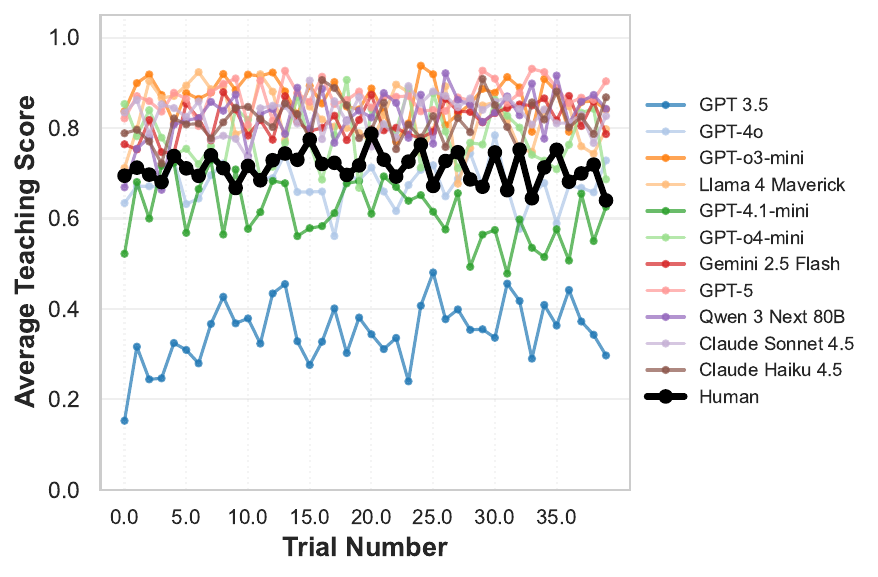}
\end{center}
\caption{\textbf{Baseline Experiment learning curves.} Mean Teaching Score across the 40 trials for humans and each LLM model. Performance is largely stable over trials, consistent with the absence of feedback.}
\label{fig:exp1_learning}
\end{figure}

\clearpage

\begin{figure}
\begin{center}
\includegraphics[width=1\textwidth]{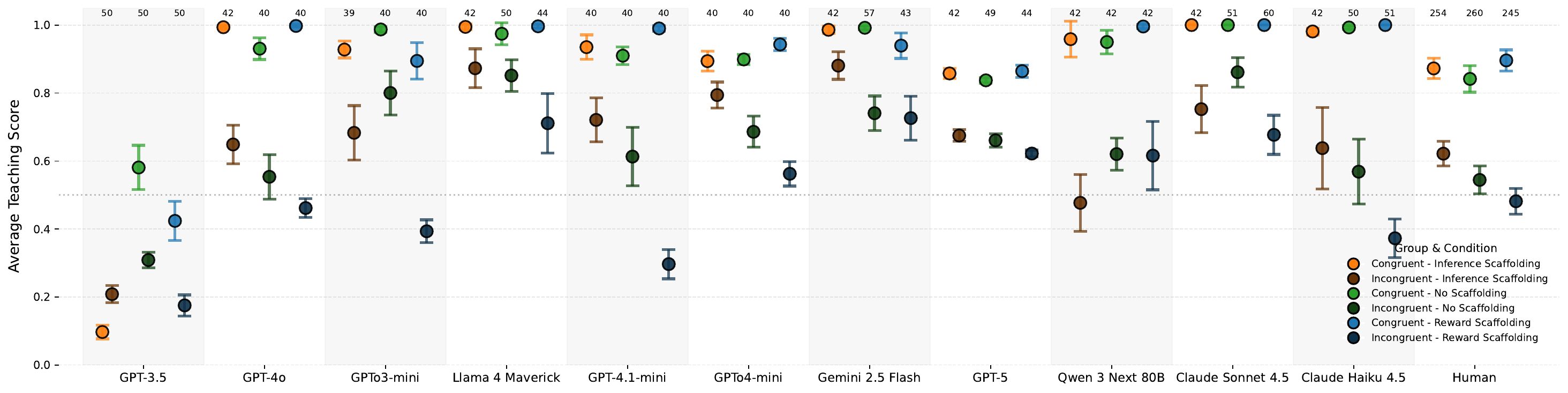}
\end{center}
\caption{\textbf{Teaching performance during training across scaffolding and congruency conditions.}
Mean Teaching Score in training for No Scaffolding, Inference Scaffolding, and Reward Scaffolding, shown separately for congruent and incongruent training groups. Error bars show 95\% CI of the mean}
\label{fig:exp3_perf_train}
\end{figure}

\clearpage

\end{document}